%% file: 2016_ICRA_MPCExp_RL_Resubmit.tex
\documentclass[letterpaper, 10 pt, conference]{ieeeconf}

\let\labelindent\relax
\input{preamble}

\input{defs}

\newcommand{\tocite}[1]{\textcolor{red}{[cite]}}
\newcommand{\colvec}[1]{\begin{bmatrix}#1\end{bmatrix}}
\newcommand{\ignore}[1]{}
\newcommand{\bb}[1]{\mathbf{#1}}

\title{\LARGE \bf
Model-based Reinforcement Learning with Parametrized Physical Models and Optimism-Driven Exploration
}
\author{%
Chris Xie \and Sachin Patil \and Teodor Moldovan \and Sergey Levine \and Pieter Abbeel%
\thanks{Department of Electrical Engineering and Computer Science, University of California, Berkeley, CA.}%
}

\begin{document}

\maketitle
\thispagestyle{empty}
\pagestyle{empty}

%%%%%%%%%%%%%%%%%%%%%%%%%%%%%%%%%%%%%%%%%%%%%%%%%%%%%%%%%%%%%%%%%%%

\begin{abstract}

In this paper, we present a robotic model-based reinforcement learning method that combines ideas from model identification and model predictive control. We use a feature-based representation of the dynamics that allows the dynamics model to be fitted with a simple least squares procedure, and the features are identified from a high-level specification of the robot's morphology, consisting of the number and connectivity structure of its links. Model predictive control is then used to choose the actions under an optimistic model of the dynamics, which produces an efficient and goal-directed exploration strategy. We present real time experimental results on standard benchmark problems involving the pendulum, cartpole, and double pendulum systems. Experiments indicate that our method is able to learn a range of benchmark tasks substantially faster than the previous best methods. To evaluate our approach on a realistic robotic control task, we also demonstrate real time control of a simulated 7 degree of freedom arm.
%We also consider the problem of controlling a simulated 7 degree of freedom arm to achieve a desired end effector position and velocity to perform a throwing task. 

%TODO: something in the abstract about being real time while previous work (Teodor) wasn't..?
\end{abstract} 

%%%%%%%%%%%%%%%%%%%%%%%%%%%%%%%%%%%%%%%%%%%%%%%%%%%%%%%%%%%%%%%%%%%
\section{Introduction}
\label{sec:intro}

%%% Comments to address from the reviews:
% 1. Clarify that control is done online
% 2. Discuss the use of general cost functions
% 3. Discuss scaling issues
% 4. Clarify how/why PILCO is more general and make it clear why we compare to it
%    - if we don't end up comparing to PILCO with prior dynamics, we should very
%      clearly explain why: we *don't* have prior dynamics, just a parameterized
%      model!
% 5. Clearly state learning times? (but they were stated already...)
% 6. Clarify that time refers to interaction time
% 7. Add baseline with true model
% 8. Discuss limitations: no model variation, no real robots
% DONE 9. Cite Riedmiller's new paper on DDP + GP for pendulum as prior work
% DONE 10. Remember to mention adaptive control methods

Model-based control of robotic systems requires model identification to be performed before the system can be effectively controlled, particularly for dynamic, high-speed motions. One way to tackle this challenge is to first perform model identification, and then use the identified model to control the system \cite{ljung1998system,hollerbach2008model}. However, this approach requires a dedicated model identification step, which can become inefficient if the dynamics change frequently or suddenly, or if the robot interacts with unfamiliar physical objects that must each be identified.

Reinforcement learning (RL) offers a framework for automatically trading off exploration and exploitation to complete the task as quickly as possible. Model-based RL reduces the required system interaction time by learning a model of the dynamics, while still trading off exploration and exploitation to learn a model that is just detailed enough to succeed at the task. General-purpose statistical models are often used to represent the dynamics, but such models can require a substantial number of samples to acquire a sufficiently accurate dynamics estimate~\cite{2011pilco,Moldovan2015}.

In this paper, we combine concepts from model identification and model-based reinforcement learning to complete the task as quickly as possible while identifying the system online to acquire a model that is sufficient for task completion. In contrast to prior methods that use generic statistical models of the dynamics, we use a feature-based least-squares formulation of the model identification problem, which allows the model to be identified extremely quickly by bringing in our prior knowledge about the robot's morphology and physics. Exploration is performed using an optimistic model-predictive control (MPC) framework~\cite{Moldovan2015}, which determines the optimal trajectory under an optimistic formulation of the system dynamics. As more interaction samples are gathered, the amount of optimism is reduced, until the method converges to the true dynamics.

\begin{figure}[t]
\centering
\includegraphics[width=0.8\linewidth]{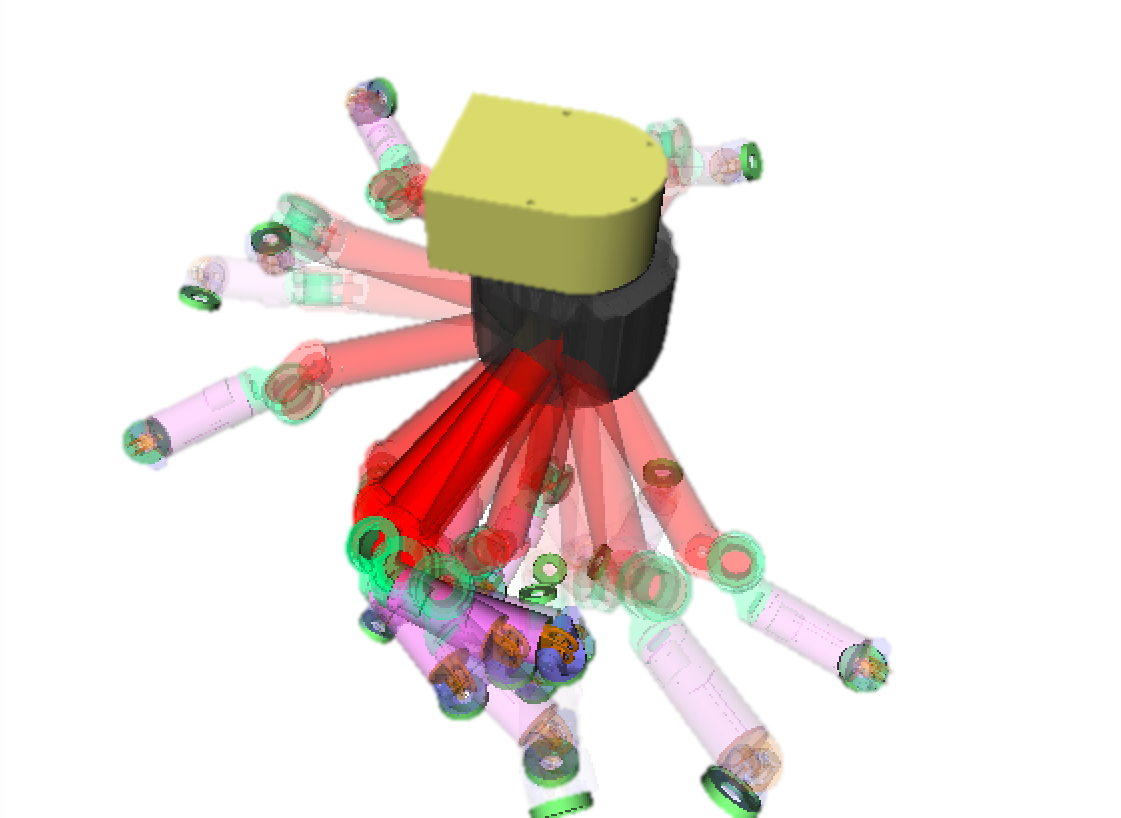}
\vspace{-5pt}
\caption{7 DoF arm learning to reach a target pose using our method. Later time steps shown with higher opacity, with the target pose fully opaque.\label{fig:teaser}}
\vspace{-10pt}
\end{figure}

Our main contribution is a method for combining optimism-driven exploration with simple least-squares model fitting based on physical features of the dynamics that can be extracted automatically from a high-level specification of the system morphology (e.g., the number and connectivity of the links, which is readily available for most robotic systems). These features can be obtained manually for simple systems, or can be computed automatically using existing packages such as SymPyBotics \cite{sousa2014physical}. We first present experimental results on the pendulum, double pendulum, and cartpole benchmarks. Compared to prior methods, our approach is able to solve these tasks using the lowest amount of system interaction time. In comparison to a prior method based on optimism-driven exploration \cite{Moldovan2015}, our approach also requires much less computation time, making it suitable for real-time online learning. To evaluate our method on a realistic robotic control task, we also demonstrate real time control of a simulated 7 degree of freedom arm, shown in Figure~\ref{fig:teaser}.

%We also consider the problem of controlling a 7 degree of freedom arm where our method is able to simultaneously identify the dynamics of the arm and achieve a desired end effector position and velocity to throw an object.

%%%%%%%%%%%%%%%%%%%%%%%%%%%%%%%%%%%%%%%%%%%%%%%%%%%%%%%%%%%%%%%%%%%

\section{Related Work}
\label{sec:background}

System and model identification has been explored extensively in the context of robotics \cite{ljung1998system,hollerbach2008model}. Several methods have been proposed in the literature for finding good excitation trajectories for model identification~\cite{swevers1997optimal,armstrong1989finding,gautier1992exciting,rackl2012robot,wang2014fast,wu2010overview}. The physical feasibility of parameters during the identification process has also been considered~\cite{gautier2013identification,sousa2014physical}. However, model identification is typically performed as a separate process from control. In contrast, our work addresses the problem of controlling a robotic system to complete a task as quickly as possible, while learning a model sufficient for task completion.

One alternative to offline model identification is provided by adaptive control \cite{Astrom:1994:AC:546778}. Adaptive control offers compelling convergence and stability guarantees, but is typically concerned with stabilizing around a target trajectory under a linear model. This makes it difficult to apply to more complex, nonlinear robotic systems with high-level goals defined by arbitrary cost functions.

Reinforcement learning (RL) \cite{kbp-rlrs-13,sutton1992reinforcement} tackles control problems with nonlinear dynamics in a more general framework, which can be either model-based or model-free. Model-based RL reduces the required interaction time by learning a model of the system during execution, and optimizing the control policy under this model, either offline in an episodic setting, or online. In the context of RL, exploration refers to intentionally taking suboptimal actions to improve future performance. Although many methods have been proposed for model-based RL with efficient exploration in discrete MDPs, data-efficient model-based RL for continuous systems remains a challenging problem despite substantial recent advances~\cite{Abbasi-yadkori2011,2011pilco,kuindersma2013variational,Moldovan2015}.

Although several very successful model-based RL methods have been proposed recently~\cite{2011pilco,boedecker14gp,Moldovan2015}, such methods typically use general-purpose statistical models of the dynamics. Such models are very flexible, but require a substantial number of samples to learn models that are accurate enough to succeed at the task. Several prior methods have suggested incorporating knowledge about the dynamics as a prior on the dynamics model~\cite{nguyen2010using,cutler14multifid,Cutler15_ICRA}. However, these methods typically assume that prior knowledge comes in the form of predictions about the next state, which is a very specific and quantitative type of prior. Our approach incorporates prior knowledge about the dynamics of rigid-body systems, but does not assume knowledge of system parameters, such as masses and link lengths. Without these parameters, this prior model cannot give reasonable predictions. However, it tells us a great deal about the \emph{structure} of the dynamics. The equations of motion can be written such that the parameters and model features decompose linearly, providing for a very efficient learning algorithm.

In order for the algorithm to learn the model at the same time as it performs the task, we use an optimism-driven exploration strategy combined with model-predictive control to continuously replan the next action. Prior work has proposed to use optimism-driven exploration mainly for discrete systems~\cite{kuindersma2013variational,ICML2012Araya_76}. A recent extension to continuous nonlinear systems provides for sample-efficient learning \cite{Moldovan2015}, but does not run in real time, making it impractical for real applications. We use an efficient MPC method based on differential dynamic programming (DDP) \cite{jacobson1970differential}, which allows us to achieve real-time performance even while continuously refitting the model parameters with each new sample.

%%%%%%%%%%%%%%%%%%%%%%%%%%%%%%%%%%%%%%%%%%%%%%%%%%%%%%%%%%%%%%%%%%%

\section{Optimistic Exploration with Continuous Model Identification}
\label{sec:approach}

Our method combines feature-based model identification with optimistic exploration in an online model-based reinforcement learning algorithm. An outline of the method is presented in Algorithm~\ref{alg:ex}. Here, $\bq$ denotes the configuration of the robot, $\bf{x} = [\dbq,\bq]$ is the state of the system, and $\bf{\tau}$ is the commanded action (e.g. the joint torques and forces). The task is specified by a cost function $l(\bf{x},\bf{\tau})$, which typically depends on the distance between the current state $\bf{x}$ and some target state $\bf{x}_{\text{goal}}$. We begin with a random initial action and empty list of observations $\mathbf{o}$. We then repeatedly execute the current action $\bf{\tau}$ for $1/\nu_c$ seconds, while collecting new observations $[\bq, \dbq, \ddbq, \bf{\tau}]$ at a frequency of $\nu_s$. These observations are appended to $\mathbf{o}$. This list of observations is used to estimate the system dynamics $\hat{f}(\bq,\dbq,\bf{\tau}) = \ddbq$, as described in Section~\ref{sec:modelid}. These estimated dynamics are then converted into optimistic dynamics $\tilde{f}(\bq,\dbq,\bf{\tau},\bm{\xi}) = \ddbq$ by including a set of \emph{virtual controls} $\bm{\xi}$ to allow the MPC algorithm to take optimistic action, as described in Section~\ref{sec:mpc_optexp}. Using these estimated dynamics, the method then plans a fixed-horizon trajectory that minimizes the cost $l(\bf{x},\bf{\tau})$ from the current state $\bf{x}$ by using differential dynamic programming (DDP), as described in Section~\ref{sec:ddp}. The next action $\bf{\tau}$ is then set to the first action along this trajectory, following the model predictive control (MPC) paradigm~\cite{camacho2013model}. This process is repeated until a specified goal condition is reached.

\begin{algorithm}[t]
\caption{Model-based RL with MPC and optimistic exploration}
\label{alg:ex}
\begin{algorithmic}[1]
\REQUIRE Start state $\bf{x}_{\text{start}} = [\dbq, \bq]_{\text{start}}$, cost function $l(\bf{x},\bf{\tau})$, sampling frequency $\nu_s$, control frequency $\nu_c$
\STATE $\bf{\tau} \leftarrow $ Random controls
\STATE $\mathbf{o} \leftarrow $ Empty list of observations
%\STATE $[\bq, \dbq]^{\intercal} \leftarrow [\bq, \dbq]^{\intercal}_{\text{start}}$
\REPEAT
\STATE Execute $\bf{\tau}$ for $1/\nu_c$ seconds
\STATE Append current $[\bq, \dbq, \ddbq, \bf{\tau}]$ to $\mathbf{o}$ every $1/\nu_s$ seconds
%\STATE $[\bq, \dbq]^{\intercal} \leftarrow $ From $[\bq, \dbq]^{\intercal}$, execute $\tau$ for $1/\nu_c$ seconds
%\STATE $\bz \leftarrow $ Append $[\bq, \dbq, \ddbq, \bf{\tau}]$ sampled every $1/\nu_s$ seconds
\STATE Estimate dynamics $\hat{f}(\bq,\dbq,\bf{\tau})$ using samples in $\mathbf{o}$
\STATE Construct optimistic dynamics $\tilde{f}(\bq,\dbq,\bf{\tau},\bm{\xi}) = \ddbq$
\STATE Optimize a trajectory from $[\dbq, \bq]$ using $\tilde{f}(\bq,\dbq,\bf{\tau},\bm{\xi})$ with virtual control penalty $\frac1m\|\bm{\xi}\|_2^2$
\STATE Update $\bf{\tau}$ to be the first action along this trajectory
\UNTIL{task completion}
\end{algorithmic}
\end{algorithm}

\subsection{Model Identification via Least Squares}
\label{sec:modelid}

In this section, we describe how the approximate dynamics $\hat{f}(\bq,\dbq,\bf{\tau})$ are fitted to samples $\mathbf{o} = \{[\bq, \dbq, \ddbq, \bf{\tau}]_i\}$. This method assumes that we know the morphology of the robot (the number and connectivity of its links), and therefore can write down its equations of motion. However, we do not necessarily know the physical parameters, such as the masses and lengths of the links. This assumption is reasonable for many physical systems, since the morphology and connectivity of the links can easily be ascertained from observation, but the physical parameters require a complex system identification procedure. For a robot consisting of a system of articulated rigid bodies, the equations of motion can be decomposed such that model identification can be formulated as linear regression, making the dynamics fitting simple and efficient. While this technique is known in the model identification literature~\cite{hollerbach2008model}, we present it here for completeness.

Using $M(\bq)$ to denote the mass matrix, $C(\bq,\dot{\bq})$ to represent the Coriolis and centripetal forces, $g(\bq)$ to represent gravity, and $\bm{\tau}$ the forces and torques, the equations of motion are given by
\begin{equation}
M(\bq)\ddbq + C(\bq, \dbq) + g(\bq) = \bm{\tau}.\label{eqn:motion}
\end{equation}
These dynamics equations can be written as:
\begin{equation}
H(\bq, \dbq, \ddbq) \cdot \Delta = \bm{\tau},\nonumber
\label{eq:dyn}
\end{equation}
where the vector $\Delta$ depends only on system parameters and the matrix $H(\bq, \dbq, \ddbq)$, also referred to as the regressor matrix, does not depend on the system parameters. Model identification can then performed by estimating $\Delta$. For instance, the vector $\Delta = \colvec{\bm{\delta}_1^\intercal& \ldots& \bm{\delta}_K^\intercal}^\intercal$ for $K$-link manipulators consists of the inertial parameters $\bm{\delta}_k$ for each link. For under-actuated systems, the dynamics can be expressed in the same form. A key difference is that we have zeroes in the $\bm{\tau}$ vector corresponding to the unactuated degrees of freedom. We describe how this decomposition is performed for specific robotic systems in Section~\ref{sec:expts}.

We assume that we have noisy observations of the features $[\bq, \dbq, \ddbq]$. Given an observation vector $\bz$ of the features $[\bq, \dbq, \ddbq]$ and generalized joint forces/torques $\bm{\tau}$ for $N$ samples, the vector $\Delta$ may be inferred by least squares regression as: 
\begin{equation}
\hat{\Delta} = \argmin_\Delta \left \|A\Delta - \mathbf{b} \right\|^2\nonumber
\end{equation}
where
\begin{equation}
A = \colvec{H(\bq_1, \dbq_1, \ddbq_1)\\ \vdots \\ H(\bq_N, \dbq_N, \ddbq_N)},\ \mathbf{b} = \colvec{\bm{\tau}_1\\ \vdots \\ \bm{\tau}_N}.\nonumber
\end{equation}
Since $H(\bq, \dbq, \ddbq)$ is typically not full rank, $A$ is not full rank and the solution is an affine subspace. Thus, we use the Moore-Penrose pseudo-inverse to get our solution $\hat{\Delta} = A^\dagger \mathbf{b}$. This gives us the least norm solution in the affine subspace. Once we estimate $\hat{\Delta}$, we can recover the forward dynamics equation $\hat{f}(\bq,\dbq,\bf{\tau}) = \ddbq$ by solving the equations of motion in (\ref{eqn:motion}) with respect to $\ddbq$. This forward dynamics estimate can then be used with any model predictive control method to choose locally optimal actions. However, acting greedily with respect to this dynamics estimate is not always desirable. When the dynamics are incorrect, it may be preferable to instead take actions that are suboptimal under the estimated model, but that are more effective at exploring the state space of the system, in order to acquire a better estimate of the dynamics that can allow the method to more quickly reach the goal. In the next section, we describe one particular exploration method that involves constructing an \emph{optimistic} estimate of the dynamics.

% Each robot link is associated with a set of inertial parameters which are comprised of mass values, center of mass coordinates with respect to the link's reference frame, and inertial tensors. Inertial tensors are symmetric and positive definite. To apply linear regression techniques, inertial parameters for the $k$th link are calculated about its frame $L_k$,
% \[
% L_k = \colvec{L_{k,xx}&L_{k,xy}&L_{k,xz}\\L_{k,yx}&L_{k,yy}&L_{k,yz}\\L_{k,zx}&L_{k,zy}&L_{k,zz}}
% \]
% which is also positive definite. Also required for linear regression is the first moment of inertia vector, $l_k$, described by
% \[
% l_k = \colvec{l_{k,x}\\l_{k,y}\\l_{k,x}} = m_k r_k =\colvec{m_k r_{k,x}\\m_k r_{k,y}\\m_k r_{k,z}}
% \]
% where $m_k$ is the mass of link $k$, and $r_k$ is the vector of center of mass coordinates with respect to the link frame.

\subsection{Optimistic Exploration}
\label{sec:mpc_optexp}

Our method uses model predictive control (MPC) to choose the actions. In order to perform the task quickly while identifying the model to a sufficient degree for task completion, we augment MPC with optimistic exploration. This combination of MPC with exploration allows for the exploration strategy to change online as the model is updated. The intuition behind this exploration strategy is that, when the dynamics are uncertain, the algorithm is allowed to choose which of the dynamics models it prefers, among those models that are highly probable given the data. If the algorithm chooses an accurate dynamics model, it will complete the task. If it chooses an inaccurate model, it will receive observations that show that this model is inaccurate, and the dynamics estimate will be improved. In prior work, this type of exploration strategy was shown to substantially improve the sample-efficiency of model-based RL~\cite{Moldovan2015}. However, this prior method suffered from very long computation times, which made it impractical for real-time online control. In this section, we present a simplified variant of the optimistic exploration framework suitable for real-time applications. In the next section, we show how it can be incorporated into a simple and efficient DDP algorithm to allow for efficient, real-time control.

In order to allow MPC to choose among the likely dynamics models, we introduce slack variables $\bm{\xi}_t$ into the dynamics, such that $\ddbq_t = \hat{f}(\bq_t,\dbq_t,\bf{\tau}_t) + \bm{\xi}_t = \tilde{f}(\bq_t,\dbq_t,\bf{\tau}_t,\bm{\xi}_t)$, where $\tilde{f}$ is the new optimistic dynamics model. When these slack variables are treated as virtual controls by MPC, they enable optimistic exploration. Intuitively, they account for uncertainty about the dynamics due to imprecise estimates of the vector $\Delta$. To keep MPC from choosing highly improbable dynamics, the slacks are penalized quadratically during MPC with a penalty of the form $\frac1m \|\bm{\xi}_t\|^2$. The magnitude of exploration is controlled by $m$, which should be proportional to the amount of uncertainty about the current dynamics.\footnote{Note that the uncertainty about the model is not the same as dynamics noise. In this work, we assume deterministic dynamics, though stochastic dynamics could also be handled in this framework.} Previous work used Bayesian models to accurately estimate this uncertainty~\cite{Moldovan2015}. In this work, we simply decrease $m$ as the number of samples $N$ increases. While this approach is somewhat simplistic, we found that it works well in practice. Establishing a formal bound on $m$ in terms of the number of samples $N$ is difficult due to the complexity of the physical model. However, we can roughly estimate this bound by considering a simplified linear-Gaussian model of the dynamics. Given a multivariate Gaussian with mean $\mu_0$ and covariance $\Sigma_0$, the variance of the posterior estimate of the mean after the update is given by $(\Sigma_0^{-1} + N\Sigma^{-1})^{-1}$, where $\Sigma$ is the sample variance~\cite{murphy2007conjugate}. This suggests that, for large $N$, the variance of the mean decreases roughly as $1/N$ with the number of samples $N$. For simplicity, we use only a single exploration hyper-parameter $c$, using $m = \frac{c}{N}$ as an estimate of the uncertainty about the model. This makes it easier to adjust the amount of exploration by tweaking a single parameter.

This optimistic exploration scheme has the effect that the system is steered into taking actions that either move it toward the goal, or else update the model if the previously chosen path to the goal is incorrect, so that another route is attempted on the next replanning step. In the case of linear dynamics or discrete systems, this optimistic exploration scheme has a number of desirable theoretical properties that make it a good choice \cite{Audibert2009,ICML2012Araya_76,AmbujTewari,Abbasi-yadkori2011}. Although such results do not exist for the general continuous nonlinear case, we observed that the optimistic exploration strategy empirically achieves effective exploration in practice. 

\subsection{Model Predictive Control}
\label{sec:ddp}

To achieve real-time control for online reinforcement learning, we use a simple and efficient differential dynamic programming (DDP) algorithm to choose locally optimal actions $\bu = [\mathbf{\tau}, \bm{\xi}]$, which include both real and virtual controls, with respect to the optimistic dynamics $\tilde{f}(\bq_t,\dbq_t,\bf{\tau}_t,\bm{\xi}_t) = \ddbq$. The actions are optimized with respect to an augmented cost function of the form \mbox{$\bar{l}(\bx,\bu) = l(\bx,\mathbf{\tau}) + \frac1m \|\bm{\xi}_t\|^2$}, which includes both the actual cost of the task and the penalty on the virtual controls. We first convert the optimistic forward dynamics into a discrete dynamics equation of the form $\bx_{t+1} = \bar{f}(\bx_t,\bu_t)$ by using a fourth order Runga-Kutta integrator, and then supply these dynamics and cost function to a DDP algorithm, which we summarize in this section for completeness. Once this algorithm determines a sequence of locally optimal actions, we extract $\bf{\tau}$ from the first action and apply this control to the system.

The optimal control problem we aim to solve can be formulated as
\begin{subequations}
\begin{align}
\min_{\bx_{1:T}, \bu_{0:T-1}} & & & \sum_{t=0}^{T-1} \bar{l}(\bx_t, \bu_t) \\
\textrm{subject to:} & & & \bx_t = \bar{f}(\bx_{t-1}, \bu_{t-1}),\ \forall t \in {1,\ldots, T}
\end{align}
\end{subequations}
The goal is to find the set of controls $\bu_{0:T-1}$ that minimizes the cost function starting from the current state $\bx_0$. We use a variant of DDP called iterative LQR (iLQR), which requires only a first order expansion of the dynamics \cite{tet-sscbo-12}. This method is particularly fast, making it well suited for MPC. The rest of this section summarizes this method. The algorithm iteratively computes first order expansions of the dynamics and second order expansions of the cost around the current trajectory, and then analytically computes the sequence of optimal controls with respect to this approximation. This sequence of controls is then executed to obtain a new trajectory, and the process repeats until convergence or for a fixed number of iterations. The controls are computed by a dynamic programming procedure that consists of recursively updating the value function and $Q$-function, defined as
\begin{align*}
V_t(\bx_t) =& \min_{\bu_{t:T-1}} \sum_{i=t}^{T-1} \bar{l}(\bx_i,\bu_i)\\
Q(\bx_t, \bu_t) =&\ \bar{l}(\bx_t, \bu_t) + V_{t+1}(\bar{f}(\bx_t, \bu_t)).
\end{align*}
Under the LQR assumptions, both of these functions are quadratic, and can be expressed up to a constant as
\begin{align*}
V_t(\bx_t) =&\ \frac12\bx_t^\intercal V_{xx,t} \bx_t + \bx_t^\intercal V_{x,t} \\
Q(\bx_t, \bu_t) =&\ \frac12 \colvec{\bx_t\\ \bu_t}^\intercal \colvec{Q_{x,t}&Q_{xu,t}\\Q_{ux,t}&Q_{uu,t}} \colvec{\bx_t\\ \bu_t} + \colvec{\bx_t\\ \bu_t}^\intercal \colvec{Q_{x,t}\\Q_{u,t}}.
\end{align*}
Let $\bar{l}_{x,t}, \bar{l}_{u,t}, \bar{l}_{xx,t}, \bar{l}_{ux,t}, \bar{l}_{uu,t}$ denote the first and second derivatives of the cost function $\bar{l}(\bx_t,\bu_t)$, and $\bar{f}_{x,t}, \bar{f}_{u,t}$ denote the derivatives of the discretized dynamics. The coefficients can be written as a recurrence described by
\begin{align*}
Q_{x,t} =&\ \bar{l}_{x,t} + \bar{f}_{x,t}^\intercal V_{x,t+1}\\
Q_{u,t} =&\ \bar{l}_{u,t} + \bar{f}_{u,t}^\intercal V_{x,t+1}\\
Q_{xx,t} =&\ \bar{l}_{xx,t} + \bar{f}_{x,t}^\intercal V_{xx,t+1} \bar{f}_{x,t}\\
Q_{uu,t} =&\ \bar{l}_{uu,t} + \bar{f}_{u,t}^\intercal V_{xx,t+1} \bar{f}_{u,t}\\
Q_{ux,t} =&\ \bar{l}_{ux,t} + \bar{f}_{u,t}^\intercal V_{xx,t+1} \bar{f}_{x,t}\\
V_{x,t} =&\ Q_{x,t} - Q_{ux,t}^\intercal Q_{uu,t}^{-1} Q_{u,t}\\
V_{xx,t} =&\ Q_{xx,t} - Q_{ux,t}^\intercal Q_{uu,t}^{-1} Q_{ux,t}.
\end{align*}
With this recurrence, we can obtain the optimal policy $g(\bx_t) = \hat{\mathbf{u}}_t + \mathbf{k}_{t} + \mathbf{K}_t(\hat{\mathbf{x}}_t - \bx_t)$, where $\mathbf{k}_t = -Q_{uu,t}^{-1}Q_{u,t}$ is the open loop term and $\mathbf{K}_t = -Q_{uu,t}^{-1}Q_{ux,t}$ is the closed loop feedback gain term. Because we are using an MPC framework, we only execute a small portion of the converged optimal control policy. This makes it very convenient to use the previous found solution as a warm-start, which allows for fast convergence.

One final detail in this framework is that, in the early stages of learning, the estimate of the model parameters $\hat{\Delta}$ may be too inaccurate to perform stable forward and backward passes with MPC. If we detect that the forward pass diverges, we revert to a simple double-integrator dynamics model. Typically, this stage of learning lasts less than one second.

%%%%%%%%%%%%%%%%%%%%%%%%%%%%%%%%%%%%%%%%%%%%%%%%%%%%%%%%%%%%%%%%%%%

\section{Experiments}
\label{sec:expts}

\begin{figure}[t]
\centering
\includegraphics[width=\linewidth]{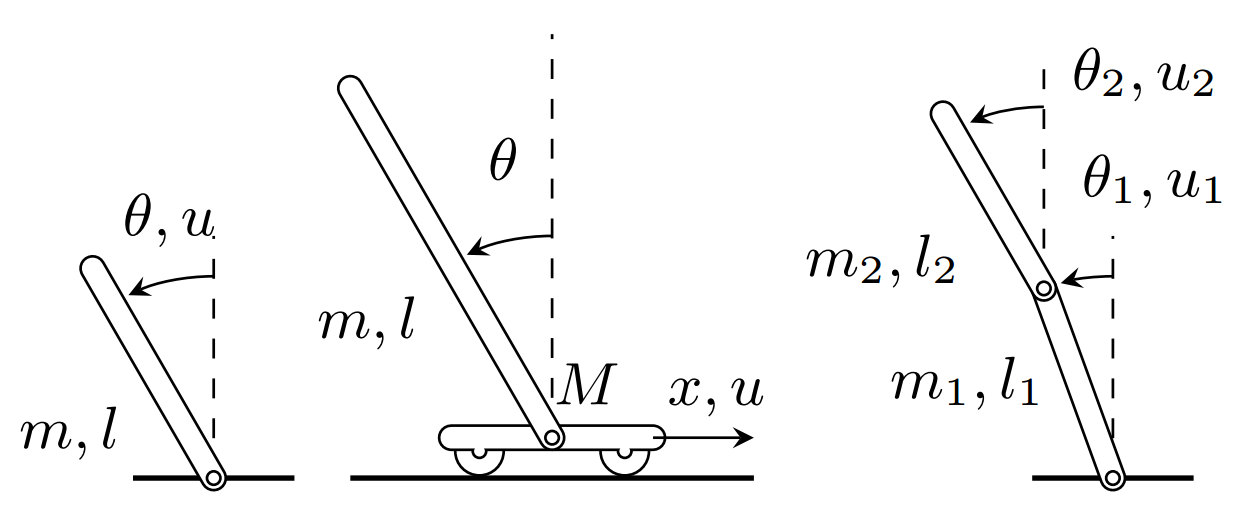}
\vspace{-5pt}
\caption{Benchmark tasks: Pendulum (left), cartpole (center), double pendulum (right)}
\label{fig:benchmarks}
\vspace{-10pt}
\end{figure}

We evaluated our method on a number of standard robotic control benchmarks: the pendulum, cartpole, and double pendulum, as shown in Figure~\ref{fig:benchmarks}, as well as on a 7 degree of freedom arm, shown in Figure~\ref{fig:sevendof}. For each system, our method obtains a noisy observation of the features $[\bq, \dbq, \ddbq]^{\intercal}$, where the noise is additive and drawn from a zero-mean Gaussian, and is tasked with reaching a target state as quickly as possible. Our implementation was in Python and C++ and ran with on a single 3.2 Ghz Intel processor.

\subsection{Benchmark Tasks}

\begin{figure*}[ht]
\centering
\begin{tabular}{r|c c c}
     & pendulum & cartpole & double pendulum \\ \hline
    DDP with known dynamics & 3.04 $\pm$ 0.89s & 7.44 $\pm$ 3.26s & 3.7 $\pm$ 0.89s\\
    our method & 3.28 $\pm$ 1.17s & 8.31 $\pm$ 3.15s & 4.98 $\pm$ 1.83s \\
    optimism-driven exploration \cite{Moldovan2015} & 3.9 $\pm$ 1s & 10 $\pm$ 3s & 17 $\pm$ 7s \\
    Boedecker et al. \cite{boedecker2014approximate} & --- & 12-18s & --- \\
    PILCO \cite{2011pilco} & 12s & 17.5s & 50s \\
\end{tabular}
\caption{Interaction time required to successfully learn each benchmark task for our method, DDP with known dynamics, and the best prior methods.}
\label{tbl:results_benchmark}
\end{figure*}

The pendulum, cartpole, and double pendulum benchmarks require controlling an underpowered system to swing up and place the endpoint of the last link at the target position. Control limits prevent each system from swinging up by continuous application of the same torque, requiring long-horizon planning. The cost function for our method consists of the distance between the endpoint of the last link and the target, as well as terms to penalize large velocities and controls. To impose control limits, we pass the controls from DDP through a squashing function of the form $s(u) = 2c\left(\sigma(u) - 0.5\right)$, where $\sigma(.)$ is the logistic function and $c$ is the control limit. The cost function therefore has the form
\begin{align*}
l(\bx_t,\bu_t) =&\ \sum_{t=0}^T \sqrt{(p(\bx_t) - \mathbf{p}^*)^\intercal Q_p (p(\bx_t) - \mathbf{p}^*) + \alpha}\\
&+ \frac12 \left [ \bx_t^\intercal Q_v \bx_t + s(\bu_t)^\intercal R s(\bu_t) + \bu_t^\intercal P \bu_t \right],
\end{align*}
\noindent where the first term is a Huber-like loss on the distance to the target endpoint position $\mathbf{p}^*$, $Q_p$ is a diagonal weight matrix, and $\alpha$ is a smoothing constant. The velocity cost is weighted by a diagonal weight matrix $Q_v$, and the controls are penalized both after squashing under $R$ and before the squashing, under $P$, as recommended in prior work~\cite{tmt-clddp-14}. Success at each task required reducing the distance between $p(\bx_t)$ and $\mathbf{p}^*$ to less than 0.05 units. We ran 50 trials for each benchmark system.

The regressor matrix for the cartpole and the double pendulum systems was obtained by manually factoring the equations of motion into $H(\bq, \dbq, \ddbq) \Delta = \mathbf{\tau}$, while the regressor matrix for the pendulum was obtained automatically using SymPyBotics \cite{sousa2014physical}. Further details about each system are presented in Appendix~\ref{app:benchmarks}.

\subsection{Benchmark Comparisons}

The results for the pendulum, cartpole, and double pendulum tasks are shown in Figure~\ref{tbl:results_benchmark}. The most sample-efficient previous results on these tasks were obtained using optimism-driven exploration with a Dirichlet process mixture model \cite{Moldovan2015}. However, the computational requirements of this approach prevented it from running in real time, with most tasks running at less than one hundredth of real time. Our proposed method is able to complete each task in real time, by using DDP-based model predictive control and a dynamics model that can be refitted efficiently using least squares. Other state-of-the-art prior methods shown in our results table include PILCO, which uses an episodic formulation instead of learning online and therefore runs comfortably in real time \cite{2011pilco}, as well as Boedecker et al.~\cite{boedecker2014approximate}, which uses Gaussian processes with MPC. We also include the time to completion for DDP using the true dynamics for each task, to provide a lower bound on the possible time to completion.

Our method achieves the best sample efficiency on each of the benchmark tasks. In fact, the time to completion on each task is very close to the time attained by DDP with known dynamics, indicating that our approach is able to identify a sufficiently accurate model of the system extremely rapidly. The advantage of our approach increases with system complexity, with the more complex double pendulum task attaining a time to completion that more than three times faster than the previous best approach \cite{Moldovan2015}, and ten times faster than the previous best approach that can run in real time \cite{2011pilco}. Furthermore, unlike the previous optimism-driven method \cite{Moldovan2015}, the computational cost of our approach is well within the bounds required for real-time operation. The average wallclock computation time for each benchmark is shown below:
%\begin{table}[h]
%\center
\begin{center}
\begin{tabular}{c c c}
pendulum & cartpole & double pendulum \\ \hline
2.67 $\pm$ 1.06s & 6.70 $\pm$ 4.49s & 3.98 $\pm$ 1.66s
\end{tabular}
\end{center}
%\caption{Wallclock computation time for each benchmark task using our method.}
%\end{table}

\subsection{7 Degree of Freedom Arm}

\begin{figure*}[ht]
\center
\begin{tabular}{r|c c c c c}
    target pose: & 1 & 2 & 3 & 4& 5 \\ \hline
    DDP with known dynamics & 1.43 $\pm$ 0.03s & 1.64 $\pm$ 0.02s & 1.34 $\pm$ 0.02s & 2.68 $\pm$ 0.84s & 1.57 $\pm$ 0.03s \\
    
    our method & 5.84 $\pm$ 2.76s & 9.11 $\pm$ 3.4s & 10.9 $\pm$ 4.62s & 9.14 $\pm$ 6.22s & 3.61 $\pm$ 1.12s  \\
    
    %Wallclock Computation & 5.49 $\pm$ 6.97s & 9.79 $\pm$ 4.4s & 10.7 $\pm$ 5.03s & 8.37 $\pm$ 6.87s & 2.94 $\pm$ 0.84s \\ \hline

    target pose: & 6 & 7 & 8 & 9& 10 \\ \hline
    DDP with known dynamics & 2.05 $\pm$ 0.0s & 0.35 $\pm$ 0.09s & 1.9 $\pm$ 0.0s & 2.65 $\pm$ 0.0s & 4.98 $\pm$ 3.32s \\
    
    our method & 6.15 $\pm$ 2.64s & 4.6 $\pm$ 2.35s & 3.71 $\pm$ 1.34s & 7.77 $\pm$ 2.36s & 9.99 $\pm$ 4.49s  \\
    
    %Wallclock Computation & 6.7 $\pm$ 3.73s & 4.89 $\pm$ 2.74s & 3.57 $\pm$ 1.9s & 6.26 $\pm$ 2.57s & 8.92 $\pm$ 4.14s \\ \hline

\end{tabular}

\caption{Results for ten randomly chosen target poses for 7 DoF arm for DDP with the true dynamics and our method, which learned the dynamics online from system interaction.}
\label{tbl:sevendof}
\end{figure*}

\begin{figure*}[t]
\centering
\includegraphics[width=\linewidth]{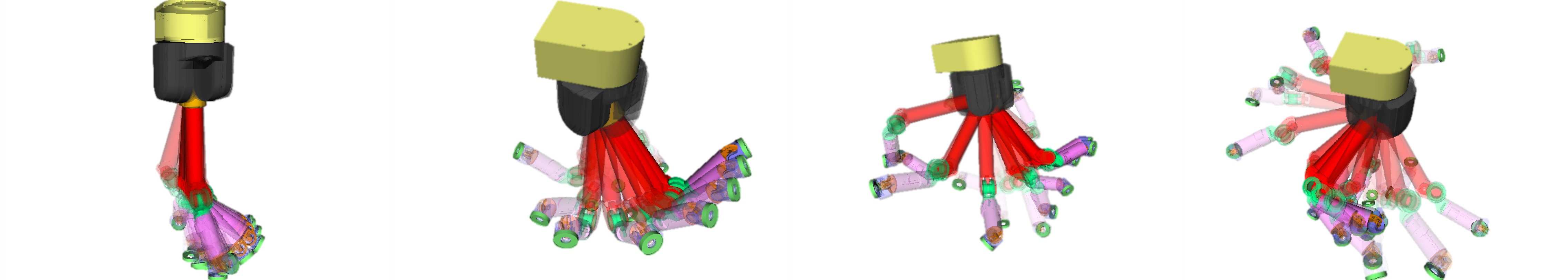}
\vspace{-5pt}
\caption{Sample trajectories for the 7 DoF Barrett arm. The target pose is opaque, and the preceding poses become progressively more translucent. Each image shows an entire trajectory executed by our method, with poses sampled at 0.25 second intervals.
\label{fig:sevendof}
}
\vspace{-10pt}
\end{figure*}

Since the dynamics features for our method can be constructed automatically, we can extend it to more complex tasks that are representative of real-world robotic control problems. To evaluate this capability, we tested our method on a simulated Barrett WAM 7 degree of freedom arm. We obtained the dynamics features by using the SymPyBotics package \cite{sousa2014physical}. The goal of the task was to reach a target pose with zero velocity, starting with no prior knowledge about the physical parameters of the system, other than the dynamics features. Ten target poses were selected at random from a spherical Gaussian distribution with a covariance of 1, centered in the middle of the joint limits. A trial was considered complete when the $L_\infty$ distance to the target pose was less than 0.05, and the velocity $L_\infty$ norm was less than 0.1. The cost function for this task had the form
\begin{align*}
l(\bx_t,\bu_t) = \frac12 \left [(\bx_t - \bx_t^*)^\intercal Q (\bx_t - \bx_t^*) + s(\bu_t)^\intercal R s(\bu_t) + \bu_t^\intercal P \bu_t \right],
\end{align*}
where $Q$ was set to be $20I$ for the velocities and $50000I$ for the joint angles. We chose $R = \text{diag}(0.08, 0.00004, 0.12, 0.04, 0.04, 0.04, 0.04)$, to account for the fact that the bigger shoulder pan joint needed to apply larger torques to raise the arm, and set $P = R/100$. We used torque limits of $[\pm 77.3, \pm160.6, \pm95.6, \pm29.4, \pm 11.6,\pm 11.6,\pm 2.7]$, and $v_c = 20$ Hz, $v_s = 100$ Hz. The observation noise was set to have a standard deviation of 0.014 by analyzing the encoder accuracy of the Barrett WAM.\footnote{As with most encoders, position readings are substantially more accurate than velocity readings. For simplicity, we set the observation noise to correspond to the less accurate velocity readings for all entries of the state.}

The results of these experiments are shown in in Figure~\ref{tbl:sevendof}. For each target pose, we ran 10 trials. The results indicate that our method was successfully able to move the arm into the desired pose in each experiment. Although some of the target poses required more time to acquire a sufficiently accurate dynamics model, some of the targets were reached in time that was only 2-3 times slower than DDP with known dynamics. In all cases, the computation time required to find a solution was comparable to the interaction time, indicating that our method could run in real time.

Note that the dynamics model of this three-dimensional 7 degree of freedom arm is much more complex than any of the benchmarks in the previous section, and the weights $\Delta$ had a dimensionality of 70, compared to less than 10 for the benchmark tasks. Several sample trajectories obtained using our method are shown in Figure~\ref{fig:sevendof}.

\section{Discussion and Future Work}
\label{sec:discussion}

We presented a model-based reinforcement learning method that combines ideas from model identification, optimistic exploration, and model predictive control to quickly and efficiently learn to perform robotic control tasks under unknown dynamics. The key idea in our work is to combine efficient linear models of the dynamics, which are informed by domain knowledge of articulated physical systems, with optimism-driven exploration. The features for these linear models can be obtained automatically from the morphology of the robot, and the optimism-driven exploration can be performed using model predictive control.

Our method achieves state-of-the-art sample efficiency on standard benchmark problems, including the pendulum, the double pendulum, and the cartpole tasks. Furthermore, unlike our previous MPC-based exploration method, which used a statistical model of the dynamics \cite{Moldovan2015}, our method achieves real-time performance, making it feasible for online reinforcement learning on real robotic platforms. Unlike the prior methods in our comparison, our approach leverages additional domain knowledge to greatly simplify the model learning problem. This prior knowledge is encapsulated in the dynamics features, which can be linearly combined to obtain the true dynamics. While this makes the comparison somewhat unequal, it serves to illustrate an important point in model-based reinforcement learning for robotic control: using freely available prior knowledge about the physical system can dramatically simplify the model learning and control problem. The prior knowledge we use is trivial to obtain for most robotic systems, since it consists of the number and connectivity of the robot's links, information that can be easily gathered from a cursory examination. %The features are then obtained automatically from this information.

%While our current results on the 7 degree of freedom arm suggest that our approach can extend to real-world robotic control problems, 
A number of future directions should be explored to make such applications effective and practical. First, although we demonstrate state-of-the-art results in simulation and evaluate some simulated unmodeled effects, we did not evaluate the robustness of our approach to unmodeled effects on a real system. The least-squares model identification procedure we use has been applied to real robotic systems in the past \cite{hollerbach2008model}, so there is reason to believe that robustness to unmodeled effects may already be adequate. However, an interesting avenue for future work would be to combine our linear models with more expressive statistical models that can account for unmodeled effects, similar to prior work on autonomous vehicle control \cite{aama-arlah-06,osb-lbnmp-14}. 

We demonstrated our method on a variety of articulated systems, but the approach is general enough to apply to other kinds of robotic systems also, including autonomous vehicles and aircraft. In fact, our prior work already demonstrated that optimism-driven exploration can achieve impressive results on simulated helicopter control, recovering from an engine failure with auto-rotation \cite{Moldovan2015}. Extension of our proposed method to such applications requires only a method for constructing the corresponding dynamics features, which can be obtained from analyzing the equations of motion of the system. Another promising application of our approach is in robotic manipulation of diverse sets of large, complex objects, where performing dedicated model identification as a discrete step is impractical. For example, if we imagine a construction robot that must handle a variety of previously unseen objects in the course of a typical manipulation scenario, the ability to very quickly acquire an accurate model for model-based planning can substantially improve the speed, efficiency, and robustness of the robot's behavior.

%This can enable a range of applications in future work.

\section*{Acknowledgements}

This research was funded in part by the NSF NRI program under award \#1227536, and by the Army Research
Office through the MAST program.

%%%%%%%%%%%%%%%%%%%%%%%%%%%%%%%%%%%%%%%%%%%%%%%%%%%%%%%%%%%%%%%%%%%

\bibliographystyle{IEEEtranS}
\bibliography{references.bib}

\appendix

\subsection{Benchmark Details}
\label{app:benchmarks}

In this appendix, we present details for each benchmark system in our evaluation.

\subsubsection{Pendulum}

The pendulum system is a simple single link system with these nonlinear dynamics \cite{2011pilco}:
\begin{subequations}
\begin{align}
\bq &= \theta, \nonumber\\
\ddbq &= \frac{u - b\dot{\theta} - \frac12 mlg \sin\theta}{\frac14 ml^2 + I}, \nonumber
\end{align}
\end{subequations}
where $\theta$ is the joint angle of the link, $u$ is the torque applied at this joint, $m$ and $l$ are the mass and length of the link, respectively, $g$ is the gravity constant $9.81$(m/s$^2$), $b$ is the friction coefficient, and $I$ is the moment of inertia around the pendulum midpoint, which is equal to $\frac{1}{12}ml^2$. The parameters are $m = 1$ kg, $l = 1$ m, \ignore{$b=0$} and constrained $u \in [-3, 3]$ N$\cdot$m. The goal state is $\colvec{0&\pi}^\intercal$, which has the pendulum standing up with no velocity.

For the cost function, we used $\alpha = 0.01,\ Q_p = 2I,\ Q_v = \mathrm{diag}([0.005, 0]),\ P = 0.01I$, where $I$ is the identity matrix. Recall that $R$ is a diagonal matrix that penalizes the squashed control and the virtual control, as mentioned before. For the squashed controls, the upper left block of $R$, which is the matrix of the quadratic penalty, is $0.01I$. We chose $T = 13, \delta = 0.1\text{s}, v_c = 10\ \text{Hz}, v_s = 100\ \text{Hz}$.

We can rewrite the dynamics in the form given in Eqn.~\ref{eq:dyn}:
\begin{subequations}
\begin{align}
H(\bq, \dbq, \ddbq) &= \colvec{\ddot{\theta}&\dot{\theta}&\sin\theta} \nonumber\\
\Delta &= \colvec{\frac13ml^2\\b\\\frac12mgl} \nonumber\\
\bm{\tau} &= u \nonumber
\end{align}
\end{subequations}

\subsubsection{Cartpole}

The cart-pole system is a nonlinear system described by the following dynamics \cite{2011pilco}: 

\begin{subequations}
\begin{align}
&\bq = [\theta~~ x]^\intercal \nonumber\\
&\ddbq = \colvec{ \frac{-3 m l \dot{\theta}^2 \sin \theta \cos \theta - 6(M + m)g\sin \theta - 6(u-b\dot{\bx})\cos \theta}{4l(M+m) - 3m l \cos^2 \theta} \nonumber \\ \frac{2m l \dot{\theta}^2\sin \theta + 3m_2 g \sin \theta \cos \theta + 4u - 4b\dot{\bx}}{4(M+m)-3m l \cos^2 \theta} }. \nonumber
\end{align}
\end{subequations}

% \[
% \ddbq = \colvec{ \frac{-3 m_2 l \dot{\theta}^2 \sin \theta \cos \theta - 6(M + m_2)g\sin \theta - 6(u-b\dot{p})\cos \theta}{4l(M+m_2) - 3m_2 l \cos^2 \theta} \\ \frac{2m_2 l \dot{\theta}^2\sin \theta + 3m_2 g \sin \theta \cos \theta + 4u - 4b\dot{p}}{4(M+m_2)-3m_2 l \cos^2 \theta} }
% \]

The state space is 4D and the control is 1D, which is the external force applied to the cart. $M$ denotes the mass of the cart, $m_2$ denotes the mass of the pole, $l$ denotes the length of the pole, $\theta$ denotes the angle of the pendulum, $p$ denotes the position of the cart, $b$ denotes the friction between the cart and the ground, and $g = 9.8$ (m/s$^2$) is acceleration due to gravity. We chose $M = .5$ kg, $m_2 = .5$ kg, $l = .5$ m, $b = .1$ N/m/s, and constrainted $u \in [-10,10]\ N\cdot m$. The goal state is $\colvec{0&0&\pi&0}^\intercal$, which has the cartpole standing up at the origin with no velocity.

For the cost function, we used $\alpha = 0.1,\ Q_p = \mathrm{diag}([1,20]),\ Q_v = \mathrm{diag}([0.07,\ 0.03,\ 0,\ 3]),\ P = 0.01I$. The upper left block of $R$, for the squashed controls, is $0.01I$. We chose $T = 8, \delta = 0.1\text{s}, v_c = 16.7\ \text{Hz}, v_s = 50\ \text{Hz}$.

Since the cartpole is underactuated, we moved a term to the right hand side of the dynamics to replace the zero due to the unactuacted degree of freedom. We can rewrite the dynamics in the form given in Eqn.~\ref{eq:dyn}:

\begin{subequations}
\begin{align}
H(\bq, \dbq, \ddbq) &= \colvec{\ddot{p}& \ddot{\theta}\cos\theta&\dot{\theta}^2\sin\theta&\dot{p}&0&0\\0&0&0&0&\ddot{p}\cos\theta&\ddot{\theta} } \nonumber\\
\Delta &= [M+m, \frac12ml, -\frac12ml, b, 3, 2l]^{\intercal} \nonumber\\
\bm{\tau} &= [u, -3g \sin\theta]^{\intercal} \nonumber
\end{align}
\end{subequations}

\subsubsection{Double Pendulum}

The double pendulum system is a fully actuated two link system with applied torques at the joints. The system dynamics are:

\begin{subequations}
\begin{align}
&\bq = [\theta_1~~ \theta_2]^\intercal \nonumber\\
&\colvec{l_1^2(\frac14m_1 +m_2) + I_1&\frac12 m_2l_2l_1 \cos(\theta_1 - \theta_2)\\ \frac12 l_1l_2m_2 \cos(\theta_1 - \theta_2) & \frac14 m_2l_2^2 + I_2}\colvec{\ddot{\theta}\\ \ddot{\theta}} = \nonumber\\
&\colvec{-l_1\left(\frac12m_2l_2\dot{\theta}_2^2\sin(\theta_1-\theta_2)-g\sin\theta_1(\frac12m_1+m_2)\right) \ignore{- b_1\dot{\theta}_1}+u_1 \nonumber\\ \frac12m_2l_2\left(l_1\dot{\theta}_1^2\sin(\theta_1-\theta_2)+g\sin\theta_2 \right) \ignore{-b_2\dbq_2}+u_2 } \nonumber
\end{align}
\end{subequations}

% \[
% \colvec{l_1^2(\frac14m_1 +m_2) + I_1&\frac12 m_2l_2l_1 \cos(\theta_1 - \theta_2)\\ \frac12 l_1l_2m_2 \cos(\theta_1 - \theta_2) & \frac14 m_2l_2^2 + I_2}\colvec{\ddot{\theta}_1\\ \ddot{\theta}_2} = \colvec{c_1\\c_2}
% \]
% \[
% \colvec{c_1\\c_2} = \colvec{-l_1\left(\frac12m_2l_2\dot{\theta}_2^2\sin(\theta_1-\theta_2)-g\sin\theta_1(\frac12m_1+m_2)\right) \ignore{- b_1\dot{\theta}_1}+u_1\\ \frac12m_2l_2\left(l_1\dot{\theta}_1^2\sin(\theta_1-\theta_2)+g\sin\theta_2 \right) \ignore{-b_2\dot{\theta}_2}+u_2 }
% \]

Here, $\theta_1$ and $\theta_2$ are the joint angles, $m_1$ and $m_2$ are the masses of link 1 and link 2, respectively. $l_1$, $l_2$ are the lengths of the links, $g$ is the gravity constant, $I_1$ and $I_2$ are the moments of inertia of the links, \ignore{$b_1,b_2$ are the friction coefficients} and $u_1,u_2$ are the torques applied at the joints. We chose $m_1=m_2=0.5$ kg and $l_1=l_2=0.5$ m. $u_1, u_2$ were constrained to be in the range $[-2,2]$ N$\cdot$m. To compute the forward dynamics, we solve this linear equation for the second derivative of the joint angles. The goal state is $\colvec{0&0&0&0}^\intercal$, which has the double pendulum standing up with no velocity.

For the cost function, we used $\alpha = 0.05,\ Q_p = 5I,\ Q_v = \mathrm{diag}([0.04,\ 0.04,\ 0,\ 0]),\ P = 0.01I$, where $I$ is the identity matrix. The upper left block of $R$, for the squashed controls, is $0.01I$. In order to make the system stabilize near the goal, we increased the control penalty to $0.1$ when the system was near the goal. We chose $T = 8, \delta = 0.08\text{s}, v_c = 16.7\ \text{Hz}, v_s = 50\ \text{Hz}$.

We can rewrite the dynamics in the form given in Eqn.~\ref{eq:dyn}:

% $$
% H(\ddbq, \dbq, \bq) = \colvec{A&\bb{0}\\ \bb{0}&B}
% $$

% $$
% A = \colvec{\ddot{\theta}_1 ~\ddot{\theta}_2\cos(\theta_1-\theta_2)~\dot{\theta}_2^2\sin(\theta_1-\theta_2)~\sin\theta_1\ignore{&\dot{\theta}_1}}
% $$
% $$
% B = \colvec{\ddot{\theta}_1\cos(\theta_1-\theta_2)~\ddot{\theta}_2~\dot{\theta}_1^2\sin(\theta_1-\theta_2)~ \sin\theta_2 \ignore{~ \ddot{\theta}_2}}
% $$
% $$
% \Delta = \colvec{l_1^2(\frac14m_1 + m_2) + I_1\\\frac12 m_2l_2l_1\\ \frac12 m_2l_2l_1 \\ -gl_1(\frac12m_1+m_2)\\\frac12 m_2l_2l_1\\\frac14m_2l_2^2+I_2\\-\frac12 m_2l_2l_1\\-\frac12m_2l_2g}
% $$
% $$
% \bm{\tau} = \colvec{u_1\\u_2}
% $$

\begin{subequations}
\begin{gather}
H(\bq, \dbq, \ddbq) = \colvec{A&\bb{0} \nonumber\\ \bb{0}&B} \nonumber\\
A = \colvec{\ddot{\theta}_1 ~\ddot{\theta}_2\cos(\theta_1-\theta_2)~\dot{\theta}_2^2\sin(\theta_1-\theta_2)~\sin\theta_1\ignore{&\dot{\theta}_1}} \nonumber\\
B = \colvec{\ddot{\theta}_1\cos(\theta_1-\theta_2)~\ddot{\theta}_2~\dot{\theta}_1^2\sin(\theta_1-\theta_2)~ \sin\theta_2 \ignore{~ \ddot{\theta}_2}} \nonumber\\
\Delta = \colvec{l_1^2(\frac14m_1 + m_2) + I_1 \nonumber\\\frac12 m_2l_2l_1 \nonumber\\ \frac12 m_2l_2l_1  \nonumber\\ -gl_1(\frac12m_1+m_2)\\\frac12 m_2l_2l_1 \nonumber\\\frac14m_2l_2^2+I_2 \nonumber\\-\frac12 m_2l_2l_1 \nonumber\\-\frac12m_2l_2g} \nonumber\\
\bm{\tau} = [u_1,u_2]^{\intercal}. \nonumber
\end{gather}
\end{subequations}

\end{document}

%% file: preamble.tex
\IEEEoverridecommandlockouts
\usepackage{url}
\makeatletter
\g@addto@macro{\UrlBreaks}{\UrlOrds}
\makeatother

\usepackage{cite}

\usepackage{times}
\usepackage{graphicx}
\DeclareGraphicsExtensions{.pdf,.png,.jpg}
\usepackage[center]{subfigure}
\usepackage{overpic}
\usepackage{color}
\usepackage[usenames,dvipsnames]{xcolor}
\usepackage{bm}
\usepackage{units}

\usepackage{makeidx}

\usepackage{mathtools}
\usepackage{amsmath, amssymb, amscd}

\usepackage{mathptmx}       % selects Times Roman as basic font
\DeclareMathAlphabet{\mathcal}{OMS}{lmsy}{m}{n}
\DeclareSymbolFont{largesymbols}{OMX}{cmex}{m}{n}

\usepackage{enumitem}

\usepackage{subfigure}
\usepackage[rightcaption]{sidecap}

\usepackage{array}

\usepackage[pdfborder={0 0 0.5}]{hyperref}
\hypersetup{
    colorlinks=true,
    linkcolor=blue,
    citecolor=blue,
    filecolor=cyan,
    urlcolor=blue
}

\usepackage{pbox}

\usepackage{algorithm}
\usepackage{algorithmic}

%% file: defs.tex
\renewcommand{\[}{\begin{equation}}
\renewcommand{\]}{\end{equation}}
\renewcommand{\b}{\backslash}

\newcommand{\bea}{\begin{eqnarray}}
\newcommand{\eea}{\end{eqnarray}}

\newcommand{\bq}{\mathbf{q}}
\newcommand{\dbq}{\dot{\mathbf{q}}}
\newcommand{\ddbq}{\ddot{\mathbf{q}}}
\newcommand{\bx}{\mathbf{x}}

\newcommand{\bu}{\mathbf{u}}
\newcommand{\bz}{\mathbf{z}}

\newcommand{\argmin}{\mathop{\mathrm{argmin}}}